# HyperNet: Self-Supervised Hyperspectral Spatial-Spectral Feature Understanding Network for Hyperspectral Change Detection

Meiqi Hu, *Graduate Student Member, IEEE,* Chen Wu, *Member, IEEE,* and Liangpei Zhang, *Fellow, IEEE*

*Abstract*—**The fast development of self-supervised learning lowers the bar learning feature representation from massive unlabeled data and has triggered a series of research on change detection of remote sensing images. Challenges in adapting self-supervised learning from natural images classification to remote sensing images change detection arise from difference between the two tasks. The learned patch-level feature representations are not satisfying for the pixel-level precise change detection. In this paper, we proposed a novel pixel-level self-supervised hyperspectral spatial-spectral understanding network (HyperNet) to accomplish pixel-wise feature representation for effective hyperspectral change detection. Concretely, not patches but the whole images are fed into the network and the multi-temporal spatial-spectral features are compared pixel by pixel. Instead of processing the two-dimensional imaging space and spectral response dimension in hybrid style, a powerful spatial-spectral attention module is put forward to explore the spatial correlation and discriminative spectral features of multi-temporal hyperspectral images (HSIs), separately. Only the positive samples at the same location of bi-temporal HSIs are created and forced to be aligned, aiming at learning the spectral difference-invariant features. Moreover, a new similarity loss function named focal cosine is proposed to solve the problem of imbalanced easy and hard positive samples comparison, where the weights of those hard samples are enlarged and highlighted to promote the network training. Six hyperspectral datasets have been adopted to test the validity and generalization of proposed HyperNet. The extensive experiments demonstrate the superiority of HyperNet over the state-of-the-art algorithms on downstream hyperspectral change detection tasks.**

## I. INTRODUCTION

CHANGE detection technology is a significant tool for urban development monitoring and land-cover/land-use change detection [1]–[3]. Generally, change detection refers to detect the changes of two remote sensing images acquired on different times at the same location [4], [5]. Hyperspectral image (HSI) provides unique spectral curve for every kind of ground object, showing great potential for object discrimination and classification [6], [7]. Hyperspectral change detection is beneficial for discriminating the changes hard to notice. The two important tasks are hyperspectral anomalous change detection (HACD) and general hyperspectral binary change detection (HBCD), respectively. HACD [8], [9] focuses on the changes that usually originate from the displacement, appearance, disappearance, and concealment of small and rare objects. HACD is motivated to highlight those rare changes used for air defense and emergency response, etc. And HBCD is targeted at detecting the precise land cover transformation under the embarrassed situation that changes may happen between some pretty similar ground objects [10], [11].

The rise of deep learning has significantly promoted the development of change detection algorithms in recent years[12]–[15]. Most of the current hyperspectral change detection methods are mainly based on the deep supervised learning[16], [17], where the training labels are necessary for the model training. The high-quality training data do provide accurate supervision for parameter training and bring about good performance, but are labor-consuming and even elusive to obtain. Thus, alleviating or eliminating the reliance of annotated labels turns into a key to the problem of hyperspectral change detection.

Self-supervised learning (SSL) [18]–[20] is a kind of new paradigm in machine learning enabling the model to learn feature representation from massive unlabeled data. And it has intrigued plenty of SSL representation learning algorithms for natural image processing. As a kind of discriminative SSL, contrastive learning (CL) [21], [22] aims at pulling together the positive sample pairs and pushing away the negative sample pairs, and has been applied for multi-spectral optical and synthetic aperture radar (SAR) remote sensing change detection[23], [24] . However, the self-supervised learning for hyperspectral change detection still faces following limitations.
1) *Pixel-level feature representation for precise change detection*. The current SSL-based change detection methods are essentially dependent on instance discrimination. Concretely, the multi-temporal images are segmented into patches for model input. The patches are mapped into feature vectors in the feature space for feature comparison. As a result, the model can only learn the patch-level feature representation, leading to coarse change detection results.
2) *Fully utilization of spatial and spectral features*. The spatial and spectral feature extraction are crucial for hyperspectral

This work was supported in part by the National Natural Science Foundation of China under Grant T2122014 and 61971317 (*Corresponding author: Chen Wu.*)

M. Hu, C. Wu and L. Zhang are with the State Key Laboratory of Information Engineering in Surveying, Mapping and Remote Sensing, Wuhan University, Wuhan 430079, China (e-mail: meiqi.hu@whu.edu.cn; chen.wu@whu.edu.cn; zlp62@whu.edu.cn).

change detection since the multi-temporal complex ground objects can be confusing due to the variant spectral difference. A common phenomenon is that the most deep-learning based change detection methods adopt two-dimensional (2D) convolutional layers to extract both spatial and spectral features simultaneously. However, with a spatial kernel filter sliding along horizontal and vertical direction, the 2D convolutional layer focuses on the spatial information along each channel but ignores the relationship between different spectral bands, leading to ineffective exploitation of spectral information.

Given the consideration of problems and concerns mentioned above, a self-supervised hyperspectral spatial-spectral feature understanding network (HyperNet) is proposed. To achieve pixel-level feature for dense prediction task like change detection, we proposed an original fully-convolutional self-supervised learning framework for convenient pixel-wise feature learning. The similarity comparisons of multi-temporal HSIs are conducted in pixel-level not the patch-level for fine-grained alignment from totally unlabeled data. And a powerful spatial-spectral attention module is designed to fully exploit the informative spatial and spectral features, including the spatial attention branch and spectral attention branch. The former one is used to explore spatial features and selectively emphasize the most informative spatial objects. The latter one is tailored to make full use of the discriminative spectral information and the relationship between local and global spectral bands. The deep spatial and spectral feature are extracted separately from the proposed two branches and then are adaptively fused to produce comprehensive features. Considering the imbalanced easy and hard samples comparison for self-supervised learning, a new loss function named focal cosine is put forward to enlarge the weight of those hard samples and promote the network training. We explore the generalization of proposed self-supervised HyperNet on two kind of change detection tasks as hyperspectral anomalous change detection (HACD) and hyperspectral binary change detection (HBCD), as SSL is known for learning general features and strong generalization for different downstream tasks. The contribution of this research can be summed up as three parts:

1) A self-supervised hyperspectral spatial-spectral feature understanding network (HyperNet) is elaborated for hyperspectral change detection. HyperNet successfully achieves pixel-level self-supervised feature representation learning for accurate dense change detection task. And a spatial attention branch and a spectral attention branch are designed to extract discriminative spatial and spectral information separately, instead of directly processing the hyperspectral images in a hybrid way.

2) In order to relieve the imbalanced problem between the easy positive and hard positive samples in SSL training, a novel similarity loss function named focal cosine loss is proposed to emphasize the weight of hard positive samples and relieve the dominance of the large majority easy samples. Experimental results tested on six hyperspectral change detection datasets indicate the effectiveness of proposed focal cosine loss function.

3) To demonstrate the validity of proposed HyperNet, we have conducted extensive experiments on hyperspectral anomalous change detection and hyperspectral binary change detection. HyperNet achieves good performance on both HACD and HBCD results and outperforms the state-of-the-art algorithms, demonstrating the effectiveness and generalization of proposed extracting spatial and spectral feature separately for better hyperspectral image understanding.

The paper is organized as follows. Section II will give a brief introduction of related works. The detailed description of the proposed HyperNet will be exhibited on Section III. The experimental results and analysis will be presented on Section IV. Finally, Section V will conclude this paper.

## II. RELATED WORK

***Hyperspectral anomalous change detection***. HACD focuses on the dynamics of small objects, facing the challenges of the violent spectral differences between the unchanged area of multi-temporal HSIs, which are mainly caused by the complex atmosphere and illumination conditions, and the motion of sensors, etc. [25], [26] The predictor-based method is a very classic solution, where the spectral values of a pair pixels of multi-temporal HSIs can be mapped by a linear or nonlinear regression model [27], [28]. And the predictive image constructed by the predictor is adjusted to be as close as the other HSI from those unchanged background areas, like the vegetation, building area. Thus, the anomalous change can be separated easily from the residual image. Chronochrome (CC) [29] presented a linear space-invariant observation model to establish a linear mapping between two HSIs, where it is assumed that the observed spectral value is in linear regression with the real primitive spectral curve. However, the inhomogeneous spectral difference filed leaves the linear model with bad performance especially under variant weather conditions. Then a novel segmented linear prediction method [30] argued to represent multi-temporal HSIs by a normal mixture model and enable the linear predictor model to differ between the background subspaces. And ACDA [31] put forward to establish an effective nonlinear mapping relationship between the multi-temporal spectral vectors by auto-encoder with multiple structural layers. It is found that these methods overemphasize the spectral information while neglect the spatial information to detect the anomalous changes.

***Hyperspectral binary change detection.*** For HBCD, one of the key problems of HBCD is how to take advantage of the detailed spectral information, where two subproblems are involved, including the refinement of the redundant adjacent spectral information and the exploration of rich spectral features separating changes from pseudo-changes. The traditional methods can be roughly categorized as three groups [32], [33], where the algebraic-based calculates the spectral difference between multi-temporal HSIs, the transformation-based compares the transformed information of new feature space, and the classification-based obtains the change detection result by comparison of the classification maps of multi-temporal HSIs. Now researchers have attempted to solve the HBCD problem with deep neural networks [34], [35]. GETNET [36] developed an end-to-end two-dimensional (2D) convolutional neural network spectral framework, where the designed mixed affinity matrix integrated with abundance maps are beneficial for full exploitation of the spectral information of HSIs. A new unsupervised spectral mapping method [37] based on auto-encoder with adversarial learning was proposed for HBCD which utilized the spectral features and minimized the reconstruction loss of input spectral vectors. [38] presented a novel strategy which combined two unsupervised model-driven method to generate credible pseudo-labels for proposed 2D HSI-CD framework. And several attention-based methods have been developed for HBCD to enhance the most informative spatial and spectral features from the huge-volume HSI data. A multilevel encoder–decoder attention network (ML-EDAN [17]) was proposed to extract hierarchical features with the designed contextual-information guided attention module. SSA-SiamNet [39] was designed to emphasize discriminative channels and locations and suppress less informative ones. However, the deep-learning-based method mostly driven by annotated samples suffer from expensive samples or even no samples in practical applications. Thus, how to fully explore spatial and spectral information from HSI data with no labels or the coarse-grained labels turns into a research priority currently[40], [41].

*Self-supervised learning*. Self-supervised feature learning is a powerful feature learning paradigm on unlabeled data [18], [19], [21], [42], and has achieved great success on visual image processing, e.g. image classification [43], video representation learning [44], target recognition [45], etc. Generally, the SSL produces massive labelled data by pre-text tasks and then learns the feature representation from the self-produced labelled data [46], [47]. And contrastive learning is one kind of discriminative SSL, aiming at attracting the positive sample pairs and push away the negative sample pairs. Data augmentation is one of the common ways to produce abundant positive and negative samples. One single image and its augmented view form a pair of positive samples, while a pair of negative samples refers to augmented views of different images. Concretely, random horizontal flip, vertical flip, random crop and resize, rotation, and noise are important and useful data transformation operators. And a great number of new methods based on siamese network have been developed to drive feature representation learning. MoCo [48] builds a large dictionary look-up to facilitates contrastive feature learning. SimCLR [49] uses plenty of negative samples within a batch to balance the feature alignment and uniformity of the samples embedded into the feature space. BYOL [50] directly removes the negative samples and predicts the view of target network by training the online network, where the target network is updated with a slow-moving average of the online network. SimSiam [51] maximizes the similarity of the prediction target of one augmented view and the embedding vector of another view from the encoder under stop-gradient constraint.

The prevailing of contrastive learning has sparked the development of advanced self-supervised learning-based change detection methods of remote sensing images[23], [24].

Currently, contrastive learning has also been applied for multi-spectral optical and synthetic aperture radar (SAR) remote sensing change detection. Under the prior probability of little changes, bi-temporal remote sensing images are regarded as the different views of the same image acquired at different time on the same location. Image patches of the same location of bi-temporal images are saw as positive sample pairs, while image patches of different location of the same training batch size are viewed as negative sample pairs. Without any labels, the network is capable of learning feature representation from input by minimizing the distance of embedding feature vectors of positive samples and pushing apart embedding feature vectors of negative samples. Generally, to avoid the collapse of the network, a large batch size is necessary for the satisfaction of enough negative samples of each batch.

Under the situation of suboptimal exploitation of spatial and spectral information and the few and limited training samples for hyperspectral change detection, the proposed method is deeply inspired by the current self-supervised learning. We proposed a novel self-supervised spatial-spectral hyperspectral understanding network to accomplish accurate change detection. The proposed method processes the hyperspectral images exactly from the perspective of two-dimensional imaging space and the spectral dimension, digging the deep and effective spatial and spectral features. Without negative sample pairs, the proposed HyperNet uses only positive sample pairs for similarity comparison. And large training batch size is not necessary for HyperNet, improving the practicability of HyperNet. Rather than patch-based sample pairs, the whole images are fed into the HyperNet to acquire global sense of information comparison and pixel-wise features for change detection.

## III. METHODOLOGY

This section gives a detailed representation of proposed approach HyperNet. Fig. 1 shows the flowchart of HyperNet for hyperspectral change detection, which is mainly composed of three parts: (a) spatial-spectral attention module, (b) projector, and (c) predictor. The whole bi-temporal hyperspectral images are firstly processed by the spatial-spectral attention module, aiming to fully exploit the spatial relationship and the diverse spectral information. The projector further maps the extracted feature embeddings to acquire deeper and more abundant features. And the predictor which is only equipped with single branch plays a role of creating the prediction target for the other branch. The self-supervised training objective is to force the projected embeddings of one view to be as similar as possible with the prediction target of the other view on the selected mask area. Under the circumstance of absence of labeled data, the pre-detection strategy provides a coarse pseudo label for supplement. A novel focal cosine similarity loss is designed to boost the contribution of those hard samples for better training. The two branches of HyperNet share the same weight, and is devoted to learn the aligned features for bi-temporal HSIs without labor-cost labels, where the distance map of the learned features is then used to separate the unchanged from the changed area easily.

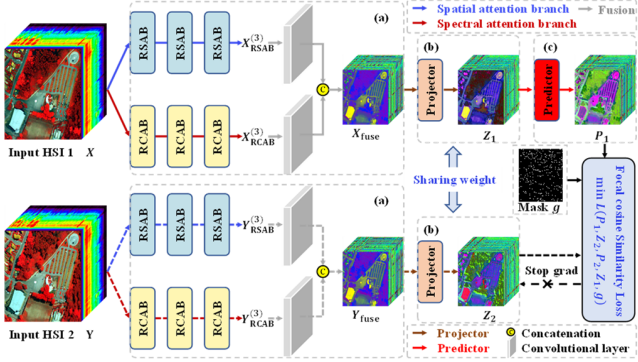

**Fig. 1** The structure of proposed HyperNet for hyperspectral change detection. (a) is the spatial-spectral attention module, which is especially designed to fully exploit the abundant information of hyperspectral images, including the residual spatial attention blocks (RSAB), residual channel attention blocks (RCAB) and the fusion block; (b) is the projector to improve the nonlinear feature mapping ability; (c) is the predictor, which is only assembled on one of branch of the HyperNet and is swapped for another branch to alternatively minimize the similarity loss.

### A. Spatial-Spectral Attention Module

The spatial-spectral attention module is elaborated to focus on the spatial and spectral features separately. The spatial attention branch concentrates on the spatial relationship from the local and global scale while the spectral attention branch contributes to explore the useful and informative spectral features. Then the spatial and spectral features are adaptively fused together to acquire the advantaged information of each pixel.

#### 1) Spatial attention branch

The spatial attention branch consists of three effective building blocks, the residual spatial attention block (RSAB), which is specially designed for spatial feature representation. The detailed architecture of RSAB is shown in Fig. 2 (a). Take the hyperspectral image $X \in \mathbb{R}^{H \times W \times C}$ as the input for example, the 3×3 2D convolutional layer followed by the Batch Normalization (BN) layer, firstly extract the shallow edge and line information within the small inceptive field. The output can be denoted as $X^{(1)} \in \mathbb{R}^{H \times W \times n}$ (Equation (1)), where $n$ refers to the number of feature maps.

$$X^{(1)} = \text{BN}(f^{3 \times 3}(X)) \quad (1)$$

For the purposed of concentration on the most informative feature, the Convolutional Block Attention Module (CBAM) [52] is incorporated after the convolution. The CBAM contains a channel-wise attention (CA) (shown in Fig. 2 (c)) and a spatial-wise attention (SA) (shown in Fig. 2 (d)), respectively. The CA squeezes the spatial information and stimulates the most discriminative spectral information from hundreds of spectral bands. Specifically, the global average and max pooling convert each channel of spatial map into a representative statistical value. Then another two shared 1×1 2D convolutional layers are targeted to gain inspired channel representations. And the summation of the extracted channel features are converted into a channel attention map $M_{\text{ca}}$ ranging from 0 to 1 by the sigmoid activation function $\sigma$ . The mathematical definition of CA can be depicted as follows:

$$M_{\text{ca}} = \sigma(f^{1 \times 1}(f^{1 \times 1}(\text{AvgPool}(X^{(1)}))) + f^{1 \times 1}(f^{1 \times 1}(\text{MaxPool}(X^{(1)})))) \quad (2)$$

where the bigger the weight is, the more attractive the spectral information of the input feature is. The stimulated channel attention feature $F_{\text{ca}}$ is obtained by elemental multiplication, which can be expressed as formula:

$$F_{\text{ca}} = M_{\text{ca}} \otimes X^{(1)} \quad (3)$$

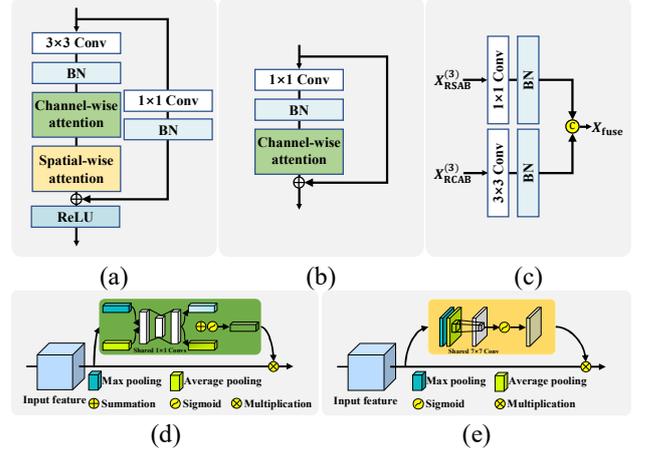

**Fig. 2** The architecture of (a) residual spatial attention block (RSAB); (b) residual channel attention block (RCAB); (c) adaptive feature fusion block; (d) channel-wise attention, and (e) spatial-wise attention.

Analogously, as shown in Fig. 2 (d), SA firstly reduces the channel number and then concentrates on how to exploit the most informative spatial information. The max pooling layer and average pooling layer of the SA focus on the edge and texture information, separately. The features extracted from the two pooling layers are further transformed by a 2D convolutional layer with a kernel filter size as 7×7. Then the sigmoid activation function transforms the features into the range from 0 to 1 to get a weight constraint map $M_{\text{sa}}$, which is finally multiplied by the input features to get the output $F_{\text{sa}}$.

$$M_{\text{sa}} = \sigma(f^{7 \times 7}(\text{AvgPool}(F_{\text{ca}}); \text{MaxPool}(F_{\text{ca}}))) \quad (4)$$

$$F_{\text{sa}} = M_{\text{sa}} \otimes F_{\text{ca}} \quad (5)$$

Noted that it is exactly the gate mechanism that controls the amount of information. The spatial information is emphasized at the location with the weight closer to 1, and suppressed at the area with the weight closer to 0.

Besides, the spatial attention block is designed in a residual architecture [53] to avoid being overfitted and convenient fusion of deep and shallow features. The input feature here is $X \in \mathbb{R}^{H \times W \times C}$, and the output feature of the spatial-wise attention is $F_{\text{sa}} \in \mathbb{R}^{H \times W \times n}$. The down-sampling on the input is necessary when the input channel is not equal to the output channel. Concretely, a 1×1 2D convolutional layer and BN are executed on the input to get $X_d \in \mathbb{R}^{H \times W \times n}$. Thus, the residual output of the first RSAB is described as $X_{\text{RSAB}}^{(1)} \in \mathbb{R}^{H \times W \times n}$:

$$X_d = \text{BN}(f^{1 \times 1}(X)) \quad (6)$$

$$X_{\text{RSAB}}^{(1)} = \text{ReLU}(X_d + F_{\text{sa}}) \quad (7)$$

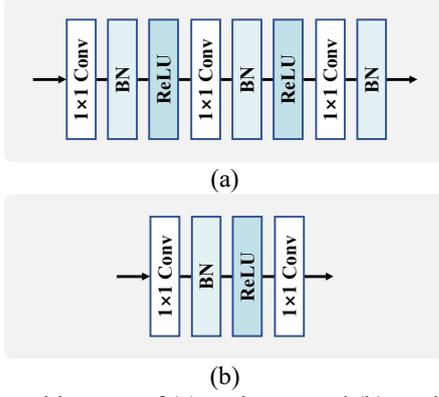

(a)

(b)

**Fig. 3** The architecture of (a) projector, and (b) predictor.

Noted that the number of channels of output feature map is set as $n$ for each RSAB block. As a result, only the first RSAB block is equipped with the down-sampling.

***2) Spectral attention branch.*** For the spectral attention branch, three residual channel attention blocks (RCAB) are specially designed for spectral feature representation. As Fig. 3 (b) shows, take the hyperspectral image $X \in \mathbb{R}^{H \times W \times C}$ as the input for example, the 2D convolutional layer with kernel size $1 \times 1$ is selected for concentrating on the spectral space. Another BN layer is used to improve the generalization of the network. The output feature map $X^{(2)} \in \mathbb{R}^{H \times W \times n}$ with the number set as $n$ can be expressed as:

$$X^{(2)} = \text{BN}(f^{1 \times 1}(X)) \quad (8)$$

The RCAB block is assembled with the channel-wise attention which is tailored to stress the most useful spectral band and get the spectral attention map $M_{\text{ca}}$ ranging from 0 to 1. The definition is defined as:

$$M_{\text{ca}} = \sigma(f^{1 \times 1}(f^{1 \times 1}(\text{AvgPool}(X^{(2)}))) + f^{1 \times 1}(f^{1 \times 1}(\text{MaxPool}(X^{(2)})))) \quad (9)$$

For the RCAB, the short-cut connection is linked between the input and the output weighted spectral feature when they share the same size. Therefore, the output of first RCAB is exactly the weighted channel attention feature, which is acquired by formula (10). And we set the number of all feature maps acquired by the RCAB as $n$.

$$X^{(1)}_{\text{RCAB}} = M_{\text{ca}} \otimes X^{(2)} \quad (10)$$

Compared with those that only a branch is designed to extract both the spatial and spectral information simultaneously, we elaborated two branches focused on spatial and spectral information, separately, for high-dimensional hyperspectral cube. The spatial attention branch stacked three RSABs to extract the high-level semantic features, while the spectral attention branch combined RCABs to extract discriminative spectral features. The spatial attention branch and the spectral attention branch are two independent branches and upgraded individually.

***3) Adaptive feature fusion block.*** After acquiring the advanced spatial features $X^{(3)}_{\text{RSAB}}$ and spectral features $X^{(3)}_{\text{RCAB}}$, an adaptive spatial-spectral feature fusion block is tailored to get the fused feature. As Fig. 2 (c) shows, we adopt the $1 \times 1$ 2D convolutional layer and BN to process the extracted spatial spectral features. The spatial feature is further enhanced and the spectral features are aggregated by the adaptive fusion block. And these two reorganized features are concatenated together alone the channel axis to get the final fused features $X_{\text{fuse}} \in \mathbb{R}^{H \times W \times (2n)}$, which can be described as follows:

$$X_{\text{fuse}} = (\text{BN}(f^{1 \times 1}(X^{(3)}_{\text{RSAB}})); \text{BN}(f^{3 \times 3}(X^{(3)}_{\text{RCAB}}))) \quad (11)$$

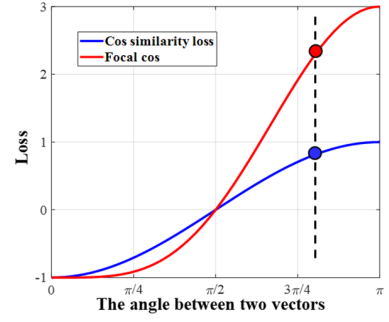

**Fig. 4** The comparison of cosine similarity loss function in blue and proposed focal cosine loss function in red.

### B. Siamese Projector and Predictor

The projector is designed to promote the extracted spatial and spectral features to get deeper embedding vectors. As presented in Fig. 3 (a), the projector is comprised of three dense $1 \times 1$ convolutional layers companied with BN layers and ReLU activation functions. Noted that these three 2D convolution layers hold the same output channel number as $n$.

The predictor with the stop-gradient operation is a key part for the self-supervised training. Only one of the branches of HyperNet is equipped with the predictor. Though the HyperNet learns to extract aligned features of bi-temporal HSIs, it may induce identical and meaningless features to force similar the output features produced by the totally siamese network. As shown in Fig. 1, the output feature of the predictor in the first branch noted as $P_1$ is forced to be similar with the output feature $Z_2$ of the projector in the second branch. Meanwhile, no gradient is back propagated in the second branch, where only the HSI $X$ is responsible for the gradient back propagation. Likewise, the output feature of the projector $Z_1$ in the first branch and the output feature $P_2$ of the predictor in the second branch are forced to be similar, where the gradient propagation is stopped in the first branch. The learned feature from predictor may differ from the one from projector. But the similarity constraint and the swap strategy facilitate the features to be more similar with each other alternatively, making the HyperNet learn meaningful aligned spatial-spectral features from the bi-temporal HSIs.

Fig. 3 (b) shows the structure of the predictor, which is composed of two dense $1 \times 1$ convolutional layers with BN as well as ReLU layer. Specifically, the first convolution layer outputs a squeezed feature map with channel number as $n$. And the final convolutional layer outputs the expanded feature map with channel number as $2n$. The projector and predictor share the same outputted feature map size.

### C. Focal Cosine Loss Function

Cosine (cos) similarity distance is a commonly used measuring index to compare the similarity of two high-dimensional vectors. The more similar the two vectors are, the closer the cosine similarity distance value is to 1. And the similarity value between the bi-temporal vectors of large difference is close to 0 or even -1. The natural imaging conditions may induce different spectral values over large areas. However, most of the spectral difference of the unchanged area are easy to be optimized. In other words, these positive samples are easy to be pulled together. Consequently, the large majority easy samples dominate during the training and the difficult samples are harder to be optimized. In order to solve the easy-sample-dominant problem, a novel similarity loss function named Focal Cosine is proposed.

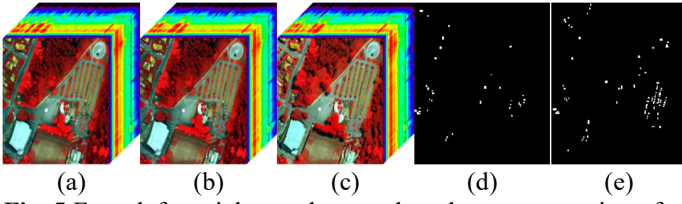

(a) (b) (c) (d) (e)

**Fig. 5** From left to right are the pseudo-color representation of "Viareggio 2013" dataset for hyperspectral anomalous change detection: (a) D1F12H1, (b) D1F12H2, (c) D2F22H2 for "Viareggio 2013" dataset, (d) reference map of EX-1: D1F12H1-D1F12H2, (e) reference map of EX-2: D1F12H1-D2F22H2.

Given $Z_1 \in \mathbb{R}^{H \times W \times (2n)}$ gained from the embedding space of first view and $P_2 \in \mathbb{R}^{H \times W \times (2n)}$ from the prediction target of the other view, the proposed loss function named as focal cosine $\mathcal{L}_{fc}$ is defined as follows:

$$\mathcal{L}_{fc}(Z_1, P_2) = -(2 - \cos(Z_1, P_2)) \otimes \cos(Z_1, P_2) \quad (12)$$

$$\text{Cos}(Z_1, P_2) = \frac{Z_1}{\|Z_1\|_2} \cdot \frac{P_2}{\|P_2\|_2} \quad (13)$$

where we call the $(2 - \cos(Z_1, P_2))$ as an adjustive factor. For the hard samples, $\cos(Z_1, P_2) \to -1$, then the $(2 - \cos(Z_1, P_2)) \to 3$. In such case, the focal cosine loss function greatly increases the weight of hard samples. As for the easy samples, $\cos(Z_1, P_2) \to 1$, $(2 - \cos(Z_1, P_2)) \to 1$. As a result, the weight of the easy samples keeps the same. As the Fig. 4 shows, the blue line refers to the original cosine similarity loss function and the red line represents the proposed focal cosine loss function. For the original loss function painted as blue, the loss achieves the minimum when the two vectors are similar enough and hold zero angles. And proposed focal cosine loss function shares the same minimum with the cosine similarity loss function. Moreover, when two vectors hold a big angle, which is obviously a pair of hard positive samples, the proposed focal cosine loss of these two vectors move up to a very high value compared with the original cosine similarity loss. When the two vectors hold a small angle, which is more a pair of easy positive samples, the proposed focal cosine loss of these two vectors drops and is lower than the value gained form the original cosine similarity loss function.

Considering extensive change may happen among multi-temporal HSIs, it heavily hampers the performance of the network to compute similarity loss for all pixels. We proposed a pre-detection strategy to get a pseudo mask $g$, where a limited number of the pixels with high probability to be unchanged are opted from the pre-detection result of a certain classic method.

The final objective function is defined as $\mathcal{L}$:

$$\mathcal{L} = \frac{1}{2} \cdot g \otimes (\mathcal{L}_{fc}(\text{StopGrad}(Z_1), P_2) + \mathcal{L}_{fc}(\text{StopGrad}(Z_2), P_1)) \quad (14)$$

The parameters of the whole network are upgraded by minimizing the loss function using gradient back propagation. And the proposed HyperNet is capable of learning powerful pixel-level deep features from the bi-temporal HSIs. Moreover, the bi-temporal features are aligned with each other in the feature space for those unchanged areas and scattered for the changed area. As a result, the learned features of the spatial-spectral attention module are quite appropriate for separating the unchanged from the changed. For the post-processing, Diff-RX [54] is used as anomaly detector to detect anomalous changes for three HBCD datasets. And the cosine similarity distance is adopted with K-means [55] is employed for thresholding for binary change maps.

## IV. EXPERIMENTS AND ANALYSIS

To test the effectiveness of proposed method on hyperspectral change detection, extensive hyperspectral experiments have been conducted on hyperspectral anomalous change detection and hyperspectral binary change detection. In this section, the descriptions of the datasets are firstly presented. Next, the details of the experimental setting are given. Then the analyses of the experimental result tested on HACD and HBCD are exhibited and the ablation experiment results are discussed.

### A. Hyperspectral Datasets

The datasets used for HACD include the benchmark dataset "Viareggio 2013" [56] and another simulated Hymap dataset.

*1) "Viareggio 2013" dataset*: Generally, the "Viareggio 2013" dataset involves three HSIs, namely D1F12H1, D1F12H2, and D2F22H2, as shown in Fig. 5. All of the three HSIs are shot in Viareggio, Italy by an airborne hyperspectral sensor SIM.GA. The spectral information covers spectrum ranging from 400 nm to 1000 nm, with a spectral resolution as 1.2 nm approximately. And the spatial resolution is 0.6 m. The image sizes are all $450 \times 375$, with 127 bands. The data are processed by de-striped [57] and spectrally binned and available at http://rsipg.dii.unipi.it/. The D1F12H1 and D1F12H2 are acquired on the same day, May 8, 2013, making up the EX-1: D1F12H1-D1F12H2. And EX-2: D1F12H1-D2F22H2 is composed of D1F12H1 and D2F22H2, which is acquired on May 9, 2013. Two HSIs of EX-1 share very similar imaging condition and the anomalous change mainly come from the movement of vehicles. But the case of EX-2 is quite different. There is severe shift between the gesture of the shooting platforms for EX-2, casting extensive distortion and great challenge for pixel-to-pixel anomalous change detection. Moreover, the weather condition of D2F22H2 has also changed compared with D1F12H1, where shadows appear around the trees and buildings.

*2) Simulated Hymap dataset*: The Simulated Hymap dataset (shown in Fig. 6) is designed to test the impact of noised on proposed method, the first image is a hyperspectral target detection dataset acquired at Cooke City, Montanta, on July 4, 2006. And it is available at http://dirsapps.cis.rit.edu/blindtest/. The image sizes are 280×800×126. The spectrum ranges cover from 453 nm to 2486 nm. To simulate different imaging condition, a noise map of random uniform distribution ranging from -10 to 10 is firstly created and then is processed by a low-pass filter with a standard deviation of ten pixels. The simulated HSI is created by an addition of the noise map and the original Hymap data. Besides, an offset of one pixel on horizontal and vertical direction are applied to the simulated HSI. And the anomalous changes are obtained by replacement of objects at other locations.

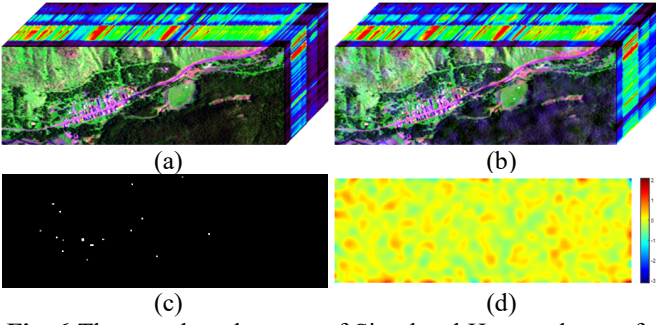

(a)            (b)

(c)            (d)

**Fig. 6** The pseudo-color map of Simulated Hymap dataset for hyperspectral anomalous change detection: (a) original Hymap image, (b) simulated Hymap image, (c) reference map, (d) the simulation noise map of random uniform distribution processed by a low-pass filter.

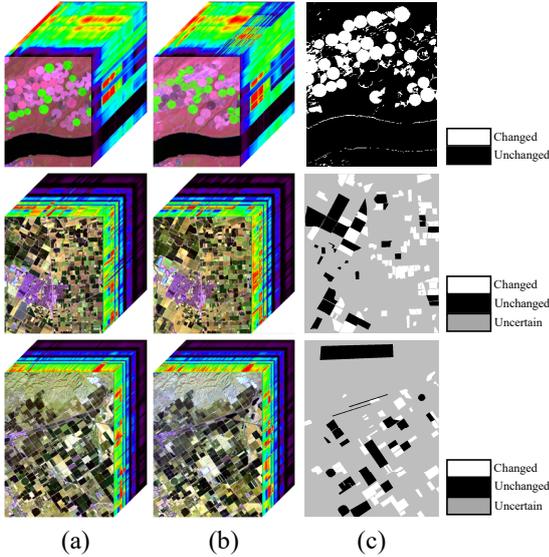

(a)       (b)       (c)

**Fig. 7** From top to bottom are pseudo-color map of Hermiston dataset, Bay dataset, and Santa Barbara dataset for hyperspectral binary change detection: (a) hyperspectral image acquired at time 1, (b) hyperspectral image acquired at time 2, (c) reference map.

Moreover, we have also tested the validity of HyperNet on another three hyperspectral binary change detection datasets.

*1) Hermiston dataset:* As the first row of Fig. 7 presents, two HSIs are acquired on May 1, 2004, and May 8, 2007 by Hyperion in Hermiston city, respectively. And Fig. 7 (c) is the reference map of HBCD. The scene covers a wide range of irrigated fields, river, cultivated land, with size as 307 × 241 pixels and 154 spectral bands.

**TABLE I**
**THE SELECTED SAMPLES FROM PRE-DETECTION RESULTS FOR SIX DATASETS**

| Dataset | Image size | Selected numbers | Ratio (%) | $n$ |
|---|---|---|---|---|
| D1F12H1-D1F12H2 | 450*375*12 | 8192 | 4.85 | 64 |
| D1F12H1-D2F22H2 | 450*375 | 8192 | 4.85 | 64 |
| Simulated Hymap | 280*800 | 8192 | 3.66 | 64 |
| Hermiston | 307*241 | 8192 | 11.07 | 72 |
| Bay | 600*500 | 8192 | 2.73 | 112 |
| Santa Barbara | 492*740 | 8192*2 | 4.50 | 112 |

*2) Bay dataset:* The second row of Fig. 7 gives three-dimensional pseudo-color cube representation of Bay dataset, which are taken on 2013 and 2015, individually, with the AVIRIS sensor surrounding the city of Patterson (California). Bay dataset is largely covered by farm lands and buildings, with spatial size as 600 × 500 pixels and 224 spectral bands. Noted that only the labeled changed and unchanged area are adopted for assessment.

*3) Santa Barbara dataset:* The last one for HBCD is exhibited as the third row of Fig. 7, where (g) and (f) are shot on the years 2013 and 2014 with the AVIRIS sensor on the Santa Barbara region. The spatial dimensions are 984 × 740 pixels and both have 224 spectral bands. The two HSIs have recorded the urban evolution and dynamic changes of farmland, and HBCD provides a powerful tool to detect the accurate dynamics changes of urban development.

### B. Implementation Details

*1) Experimental Settings:* We implemented our method by Pytorch and conducted experiments on a single NVIDIA RTX 3090 GPU. And the parameters of the network are initialized by He-normal way [58]. The optimizer is SGD with momentum as 0.9 and L2 normalization efficient as 0.0001. We set the initial learning rate as 0.05 and a cosine decay [59] strategy is adopted. The number of total epochs for training is set 200. Since the proposed method is a full-convolutional neural network, the whole image is fed into the network without the need of batch size. As for the pre-detection strategy, Diff-RX is opted for pre-detection methods for HACD and Change Vector Analysis (CVA) is opted for HBCD. The pseudo samples are selected from the pre-detection result. The selected numbers of pseudo samples for all datasets are listed as TABLE I. Noted that Santa Barbara dataset are segmented in two owing to the large size of 984×740 for HyperNet. The parameter settings of feature num $n$ are summarized in TABLE I.

*2) Comparison Methods:* To testify the effectiveness of proposed method, several methods are opted for comparison. For HACD, another eight methods, namely, Chronochrome (CC) [29], Unsupervised Slow Feature Analysis (USFA) [60], Diff-RX [54], Hyperbolic Anomalous Change Detector (HACD) [9], Simple Difference Hyperbolic Anomalous Change Detector (SDHACD) [9], Straight Anomalous Change Detector (SACD) [9], MTC-NET (patch=13), and MTC-NET

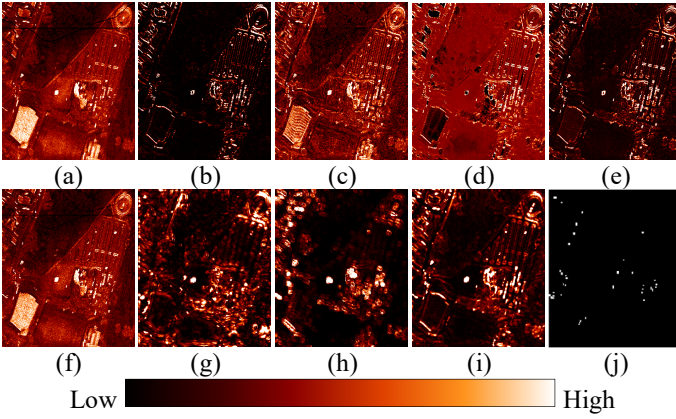

| | | | | |
|---|---|---|---|---|
| (a) | (b) | (c) | (d) | (e) |
| (f) | (g) | (h) | (i) | (j) |

Low ▬▬▬▬▬▬▬▬ High

**Fig. 8** The anomalous change detection maps of EX-1: D1F12H1-D1F12H2 on (a) CC, (b) USFA, (c) Diff_RX, (d) HACD, (e) SDHACD, (f) SDACD, (g) MTC-NET (patch=13), (h) MTC-NET (patch=31), (i) proposed HyperNet, (j) Reference change map.

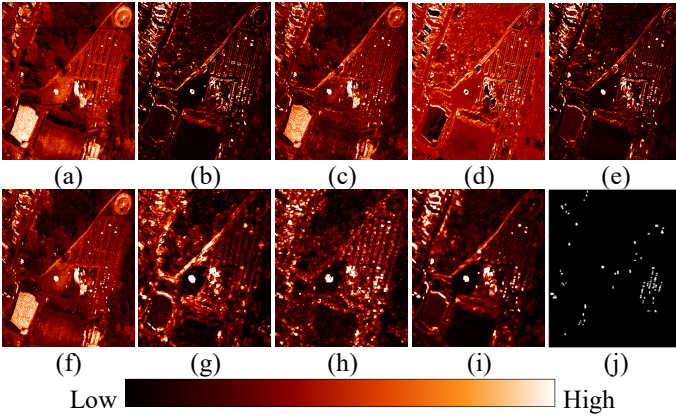

| | | | | |
|---|---|---|---|---|
| (a) | (b) | (c) | (d) | (e) |
| (f) | (g) | (h) | (i) | (j) |

Low ▬▬▬▬▬▬▬▬ High

**Fig. 9** The anomalous change detection maps of EX-2: D1F12H1-D2F22H2 on (a) CC, (b) USFA, (c) Diff_RX, (d) HACD, (e) SDHACD, (f) SDACD, (g) MTC-NET (patch=13), (h) MTC-NET (patch=31), (i) proposed HyperNet, (j) Reference change map.

(patch=31) [61]. In order to stress the effect of the patch size on the performance of change detection, MTC-NET (patch=13) with patch size set as 13×13, MTC-NET (patch=31) with patch size set as 31×31 are calculated for comparison. For HBCD, another six algorithms, CVA, Iterative Slow Feature Analysis (ISFA) [62], MSCD [24], OSCD [23], MTC-NET (patch=13), and MTC-NET (patch=31) are selected for comparison. Noted that OSCD and MSCD are two patch-based self-supervised learning method for multi-sensor change detection, and the latter has been tested on multi-spectral image change detection. The batch size of MSCD is set as 64 according to the [24], while the OSCD adopts patch size as 16 for Bay and Santa Barbara datasets, and 8 for Hermiston dataset. Except for OSCD, the rest comparative deep-learning-based methods all set the batch size as 256. And OSCD employs the multi-view contrastive loss of one positive sample and another N-1 negative sample within a simple batch, where the batch size is set N. Therefore, the batch size of OSCD is set as 1024.

*3) Evaluation Criteria*: Moreover, the Receiver Operating Characteristic (ROC) and Area under Curve (AUC) are computed for quantitative evaluation of HACD, while OA, Kappa, F1 score, Precision and Recall rate are calculated for quantitative assessment of HBCD. The detection maps are segmented into four parts, True Positive (TP), True Negative (TN), False Negative (FN) and False Positive (FP).

### C. Hyperspectral Anomalous Change Detection Results

The anomalous change detection results of eight comparative methods and proposed HyperNet on EX-1: D1F12H1-D1F12H2 are represented in Fig. 8. The darker the detection map is, the more likely the area is unchanged. And vice versa. It can be found that the background of USFA, SDHACD, MTC-NET (patch=31), and proposed HyperNet are nearly in black, indicating good performance on the compression of the background. However, for MTC-NET (patch=31), the large input patch size induces the wide range of information sharing and the decrease of location accuracy. As a result, there is some omission of anomalous changes in the detection map of MTC-NET (patch=31). Although a smaller input patch size reduces those effect, much more noise appears in the detection map of the MTC-NET (patch=13). Based on fully convolutional neural network, the proposed HyperNet balances the need of spatial information and location accuracy with the design of spatial and spectral attention module, where the local spatial information, pixel-level spectral information, and the attention from global scale are integrated. As Fig. 8 (i) shows, the anomalous changes are almost highlighted and the edges of the change area are preserved in the detection map of HyperNet.

Fig. 9 shows the anomalous detection maps of all methods on EX-2: D1F12H1-D2F22H2. Owning to the violent shift of the sensor, obvious brightness can be observed at the edge of the pine trees area in all results, extending from the top right to left middle. Most of the background of the results of USFA, HACD, and proposed HyperNet are at low value, showing nice performance on reducing the impact of pervasive spectral difference. Lots of noise can be observed on the results of MTC-NET (patch=13) and MTC-NET (patch=31). For EX-2, the anomalous changes mainly hail from the appearance and disappearance of the vehicles, and are concentrated in the right center, which is a parking lot. Compared with the reference change map, HACD SDHACD, and HyperNet are capable of detecting nearly all anomalous changes, where the anomalous change values of the result of HyperNet seem to be brighter than the rest. In contrast with the maps of MTC-NET (patch=13) and MTC-NET (patch=31), there is no sharp edge expansion in the change region and the omission and noise is less on the result of HyperNet, indicating the better performance of proposed pixel-level HyperNet. The two-branch spatial-spectral attention module are tailored to focus on spectral and spatial information, separately, where sufficient features of two distinct dimensions are abstracted.

The anomalous change detection results of all methods on the Simulated Hymap dataset are presented in Fig. 10. Owing to the offset on both horizontal and vertical direction, there are massive misplacements on the center of the image, which is a dense mixture of buildings, vehicles, and roads. All methods gain high values on the center area except for the HACD. Moreover, the noise also has a negative impact on detecting the anomalous changes. Obvious noise speckles can be found in the detection maps of all comparative methods except USFA. And the noise impact is magnified especially in the detection map of MTC-NET (patch=31). However, the noise has little influence on the detection result of HyperNet, indicating strong noise resistance of proposed method. Besides, HyperNet detects most simulated anomalous changes, gaining good performance on anomalous change detection.

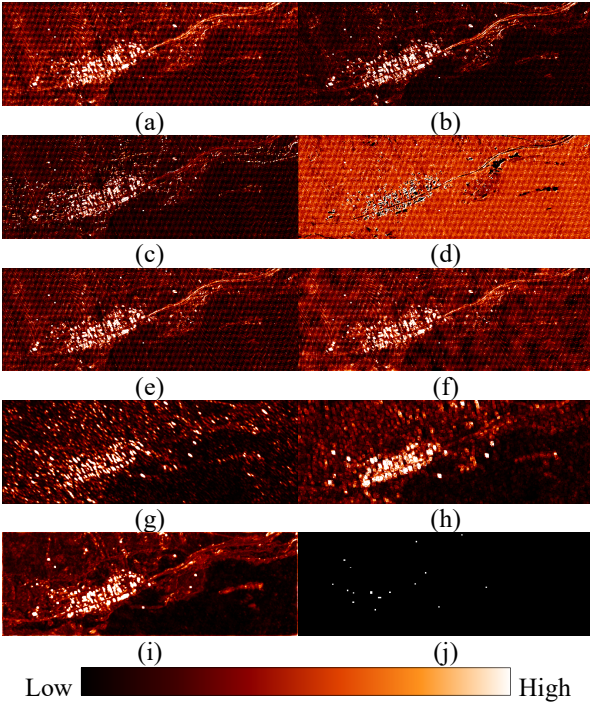

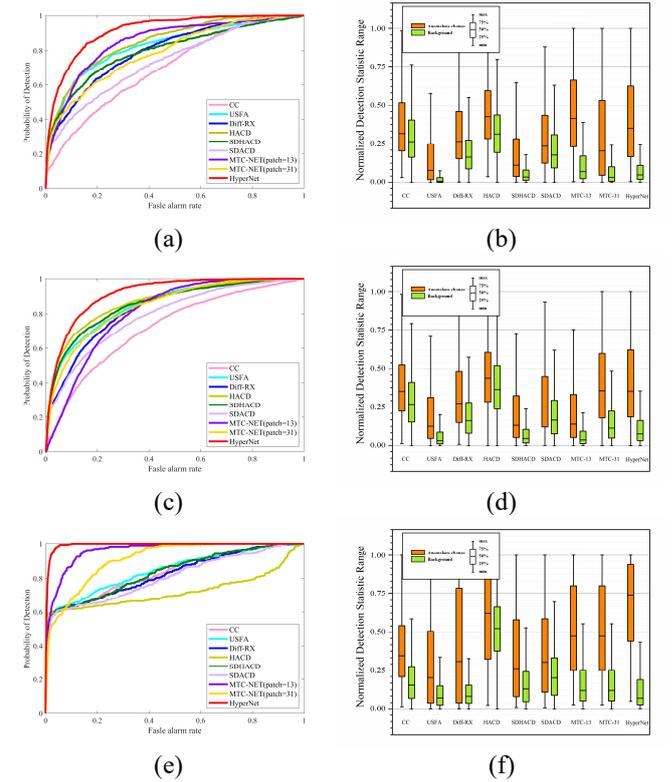

**Fig. 10** The anomalous change detection maps of Simulated Hymap dataset on (a) CC, (b) USFA, (c) Diff_RX, (d) HACD, (e) SDHACD, (f) SDACD, (g) MTC-NET (patch=13), (h) MTC-NET (patch=31), (i) proposed HyperNet, (j) Reference change map.

**Fig. 11** The first column are the ROC curves of all methods on (a) EX-1: D1F12H1-D1F12H2, (c) EX-2: D1F12H1-D2F22H2, (e) Simulated Hymap dataset, and the second column are separability maps of all methods on (b) EX-1: D1F12H1-D1F12H2, (d) EX-2: D1F12H1-D2F22H2, (f) Simulated Hymap dataset.

Quantitative assessment is also provided to test the validity of proposed method. The first column of Fig. 11 shows the ROC curves of all methods on three datasets, where the horizontal axis refers to the false alarm rate and the vertical axis is the probability of detection, respectively. And the method that gains high probability of detection at low false alarm rate is a terrific anomalous change detector. The ROC curve of proposed HyperNet in red overtops the rest comparative methods for all three datasets. And the MTC-NET (patch=13) in purple obtains good performance on the EX-1: D1F12H1-D1F12H2 and Simulated Hymap dataset; HACD in light green also acquires comparative effect on the EX-1: D1F12H1-D1F12H2 and EX-2: D1F12H1-D2F22H2.

The second column of Fig. 11 are the separability maps of all methods on three datasets. The output value of anomalous changes and background are collected, respectively, to test the ability of separating anomalous change and background. The values are normalized to [0, 1] for convenience of visualization. The box in orange refers to the anomalous change, while the box in green denotes the background. The top and bottom of each box represent the maximum and minimum, individually, and the main part of the box hold values form the smallest 25% to the largest 75%. For EX-1: D1F12H1-D1F12H2, USFA, SDHACD, and MTC-NET (patch=31) can suppress the background values at low range, but their separability with the anomalous change values are less perfect. And there is larger intersection between the anomalous change box and background box of CC, Diff-RX, HACD and SDACD. By contrast, MTC-NET (patch=13) and HyperNet gain large gap between corresponding boxes, where the values of background of HyperNet fall into a lower range. The situation of EX-2: D1F12H1-D2F22H2 is very similar with EX-1. However, MTC-NET (patch=13) and MTC-NET (patch=31) both acquire larger background values, which intersect with corresponding anomalous change values to some extent. HyperNet gains good separability between the anomalous change box and background box. As for the Simulated Hymap dataset, there is a large distance separating the anomalous change from the background for the result of HyperNet. It is observed that the result of USFA and Diff-RX hold low anomalous change values, but they are intersected with the anomalous change values set to a great degree. In addition, HACD gains low separability, which induces a dropped ROC curve shown in Fig. 11 (e). Although it is a simulated dataset, the noise from a random uniform distribution processed with low-pass filtering and the offset on whole image cast great challenge for anomalous change detection. HyperNet obtains good performance discriminating the anomalous change with the complex background. TABLE II gives a comparison of AUC performance of all method on three datasets. The AUC means the area under the ROC value, and is the comprehensive index for an anomalous change detector. The highest is in bold and the second best is underlined. For all three datasets, HyperNet outperforms other methods with best AUC of 0.9147 for EX-1, 0.9126 for EX-2, and 0.9927 for Hymap dataset, which is much higher than the second best. HACD ranks second with AUC as 0.8489 for EX-1: D1F12H1-D1F12H2 and 0.8666 for EX-2: D1F12H1-D2F22H2. MTC-NET (patch=13) gains second best for Simulated Hymap dataset with AUC equal to 0.9589.

TABLE II
THE AUC COMPARISON OF ALL METHODS ON THREE
HYPERSPECTRAL ANOMALOUS CHANGE DETECTION DATASETS

| Method | EX-1 | EX-2 | Simulated Hymap |
|---|---|---|---|
| CC | 0.6991 | 0.7196 | 0.8327 |
| USFA | 0.8220 | 0.8479 | 0.8437 |
| Diff-RX | 0.7993 | 0.8277 | 0.8249 |
| HACD | 0.8489 | 0.8666 | 0.7232 |
| SDHACD | 0.7978 | 0.8525 | 0.8343 |
| SDACD | 0.7372 | 0.7896 | 0.8069 |
| MTC-NET (patch=13) | 0.8346 | 0.8019 | 0.9589 |
| MTC-NET (patch=31) | 0.7776 | 0.8411 | 0.9276 |
| HyperNet | 0.9147 | 0.9126 | 0.9927 |

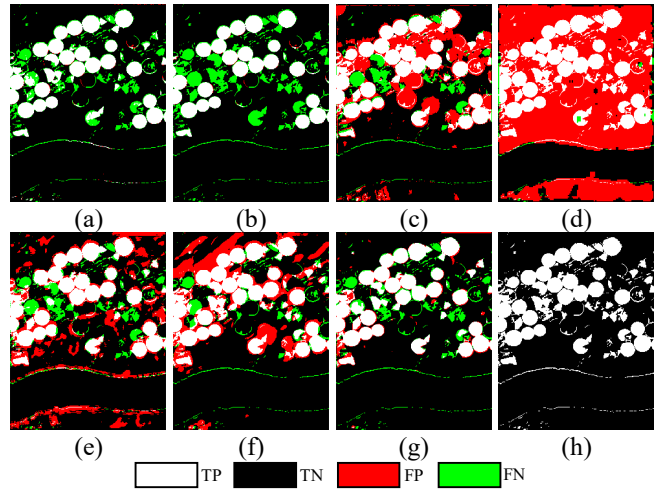

**Fig. 12** The binary change detection maps on Hermiston dataset on of (a) CVA, (b) ISFA, (c) MSCD, (d) OSCD, (e) MTC-NET (patch=13), (f) MTC-NET (patch=31), (g) proposed HyperNet, (h) Reference change map.

### D. Hyperspectral Binary Change Detection Results

The binary change detection results of all comparative methods and proposed HyperNet on the Hermiston dataset is showed in Fig. 12. Concretely, the TP is in white; TN is in black; FP is in red, and FN is in green. There are massive false detections in red in the detection map of the OSCD (Fig. 12 (d)), which is consistent with the low Precision as 0.2802 presented in TABLE III. It is analyzed that large batch size is necessary for the training of OSCD while the Hermiston dataset is a small size image with 307 × 241. And there are also some red false detections in the results of MSCD, MTC-NET (patch=13), and MTC-NET (patch=31), especially in the left bottom corner, where some buildings are located. And some missing detection in green can be found in the results of CVA and ISFA. By contrast, HyperNet is capable of detecting most of the changes with few false detections. TABLE III lists the quantitative assessment of all method on Hermiston dataset. The maximum is in bold and the second largest in underlined. CVA gains the best OA and Kappa as 0.9272 and 0.7670, separately. HyperNet obtains the comparative OA as 0.9206 and Kappa as 0.7613. The highest F1 score is 0.8112 acquired by HyperNet. Besides, the ISFA gains the largest Precision and the CVA ranks second, indicating the detected changes are almost really changed area.

TABLE III
THE QUANTITATIVE EVALUATION OF ALL METHODS ON
HERMINTON DATASET

| Method | OA | Kappa | F1 | Precision | Recall |
|---|---|---|---|---|---|
| CVA | **0.9272** | **0.7670** | 0.8103 | 0.9819 | 0.6898 |
| ISFA | 0.9023 | 0.6716 | 0.7262 | **0.9852** | 0.5750 |
| MSCD | 0.7851 | 0.4788 | 0.6201 | 0.5154 | 0.7782 |
| OSCD | 0.4332 | 0.1307 | 0.4344 | 0.2802 | **0.9656** |
| MTC-NET (patch size=13) | 0.7960 | 0.4741 | 0.6011 | 0.6111 | 0.6581 |
| MTC-NET (patch size=31) | 0.8343 | 0.5399 | 0.6468 | 0.6370 | 0.6733 |
| HyperNet | 0.9206 | 0.7613 | **0.8112** | 0.8740 | 0.7569 |

The OSCD detects all the changes with the cost of a low Kappa and Precision. On the whole, HyperNet gets the best comprehensive performance on Hermiston dataset. The proposed double-branches spatial-spectral attention module are targeted at extracting spatial objects relation and the discriminative spectral features, beneficial to reduce the false detection and improve the change detection, making a difference in the change detection result of Hermiston dataset.

Fig. 13 represents the binary change detection results on Bay dataset. The uncertain area which is unlabeled is in gray. MSCD and OSCD gain a little more omitted changes in green than other methods especially for the left farmland area. And these two methods are originally designed for multi-spectral remote sensing change detection, facing challenges dealing with the hyperspectral images. Compared with the MTC-NET (patch=31), more false detections in red can be found in the detection map of MTC-NET (patch=13). However, MTC-NET (patch=31) omits more changes in green than MTC-NET (patch=13) does. The selection of batch size plays an import role in the detection performance. By contrast, HyperNet with a patch-free architecture avoids this problem and gains better change detection result with less false alarms. TABLE IV shows the quantitative evaluation results of Bay dataset. HyperNet acquires the best OA as 0.9079, Kappa as 0.8152, F1 score as 0.9129 and Recall rate as 0.9037 among all method, and outperforms the second largest one. And ISFA obtains the highest Precision as 0.9695 with lest false detection.

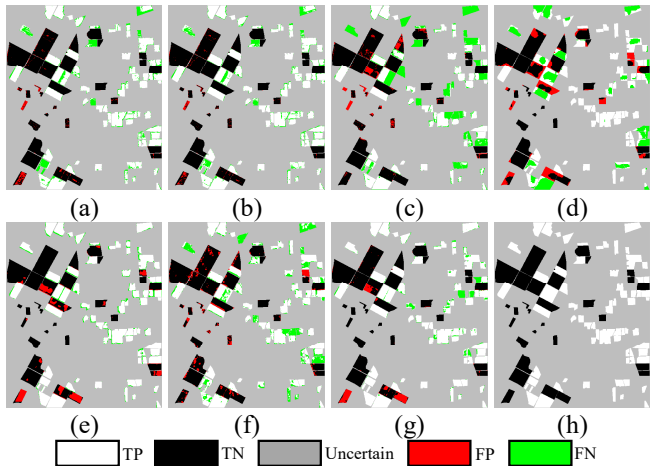

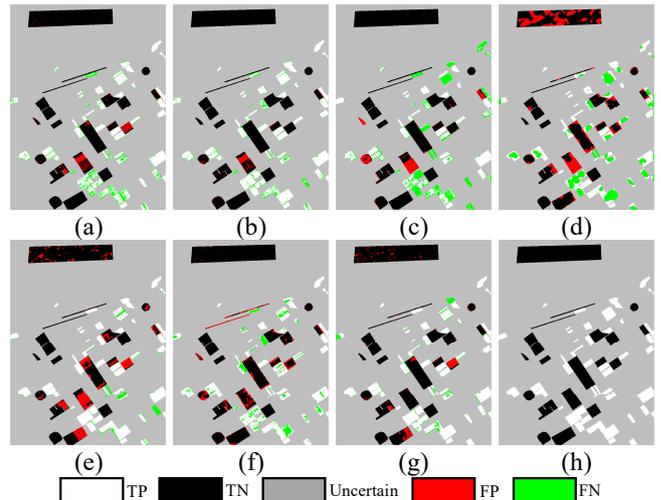

**Fig. 13** The binary change detection maps on Bay dataset on of (a) CVA, (b) ISFA, (c) MSCD, (d) OSCD, (e) MTC-NET (patch=13), (f) MTC-NET (patch=31), (g) proposed HyperNet, (h) Reference change map.

**Fig. 14** The binary change detection maps on Santa Barbara dataset on of (a) CVA, (b) ISFA, (c) MSCD, (d) OSCD, (e) MTC-NET (patch=13), (f) MTC-NET (patch=31), (g) proposed HyperNet, (h) Reference change map.

TABLE IV
THE QUANTITATIVE EVALUATION OF ALL METHODS ON BAY DATASET

| Method | OA | Kappa | F1 | Precision | Recall |
|---|---|---|---|---|---|
| CVA | 0.8723 | 0.7462 | 0.8708 | <u>0.9474</u> | 0.8057 |
| ISFA | <u>0.8917</u> | <u>0.7848</u> | 0.8905 | **0.9695** | 0.8234 |
| MSCD | 0.7503 | 0.5089 | 0.7260 | 0.8777 | 0.6190 |
| OSCD | 0.7264 | 0.4533 | 0.7339 | 0.7644 | 0.7065 |
| MTC-NET (patch size=13) | 0.8821 | 0.7627 | <u>0.8910</u> | 0.8807 | <u>0.9016</u> |
| MTC-NET (patch size=31) | 0.8108 | 0.6240 | 0.8082 | 0.8818 | 0.7459 |
| HyperNet | **0.9079** | **0.8152** | **0.9129** | 0.9224 | **0.9037** |

TABLE V
THE QUANTITATIVE EVALUATION OF ALL METHODS ON SANTA BARBARA DATASET

| Method | OA | Kappa | F1 | Precision | Recall |
|---|---|---|---|---|---|
| CVA | 0.8780 | 0.7403 | 0.8376 | 0.8792 | 0.7997 |
| ISFA | <u>0.8912</u> | <u>0.7675</u> | 0.8535 | **0.9071** | 0.8059 |
| MSCD | 0.7868 | 0.5313 | 0.6872 | 0.8127 | 0.5952 |
| OSCD | 0.6869 | 0.3384 | 0.5914 | 0.6137 | 0.5751 |
| MTC-NET (patch size=13) | 0.8805 | 0.7550 | <u>0.8566</u> | 0.8288 | <u>0.8920</u> |
| MTC-NET (patch size=31) | 0.8886 | 0.7641 | 0.8539 | 0.8801 | 0.8301 |
| HyperNet | **0.9114** | **0.8148** | **0.8880** | 0.8836 | **0.8925** |

The binary change detection results of Santa Barbara dataset are showed in Fig. 14. It is observed that plenty of false detections in red appears in the detection maps of OSCD and MTC-NET (patch=13), corresponding to low Precision shown in TABLE V. And obvious missing detection can be found in the results of CVA, MSCD, and OSCD, consistent with low recall rate shown in TABLE V. HyperNet is able to detect most of the changes under low false detection rate. As TABLE V presents, HyperNet gains the best OA as 0.9114, Kappa as 0.8148, F1 score as 0.8880, and Recall as 0.8925 among all comparative methods, while the ISFA gain the highest Precision with 0.9071. To conclude, HyperNet is able to detect more precise change detection result, capturing subtle changes of the farmland in a self-supervised learning way.

### E. The Ablation Experiments of HyperNet

To test the performance of proposed spatial-spectral attention module and the focal cosine loss function on change detection,

extensive ablation experiments have been designed. Concretely, the "Base" is designed using basic normal two-dimensional convolutional layers with the cosine similarity function. And "Base + SSA" denotes the spatial-spectral attention module is adopted compared with "Base".

TABLE VI gives AUC performance of ablation experiments of HyperNet on three hyperspectral anomalous change detection datasets. The largest one is bold and the second largest is underlined. As shown in TABLE VI, the "Base + SSA" gains better AUC than "Base" does for all three datasets. And HyperNet acquires higher AUC than "Base + SSA" does. For intuitional representation of the validity of focal cosine loss function, Fig. 15 exhibits the ROC comparison of the ablation experiments for three datasets. The "Base + SSA" in blue gains higher probability of detection than "Base" in black does. And more anomalous changes are highlighted in the results of "Base + SSA" (shown in the second row of Fig. 16) than the maps of "Base" (shown in the first row of Fig. 16), representing the effectiveness of the proposed spatial-spectral attention module. The ROC curve of HyperNet in red outperforms the "Base+ SSA" in blue for all three datasets (shown in Fig. 15), where the focal cosine loss function is encouraged to put

TABLE VI
THE AUC COMPARISON OF HYPERNET WITH AND
WITHOUT PROPOSED FOCAL COSINE LOSS ON THREE
HYPERSEPCTRAL ANOMALOUS CHANGE DETECTION DATASETS

| Method | EX-1 | EX-2 | Simulated Hymap |
|---|---|---|---|
| Base | 0.8562 | 0.8966 | 0.9886 |
| Base + SSA | 0.9081 | 0.9112 | 0.9923 |
| HyperNet | **0.9147** | **0.9126** | **0.9927** |

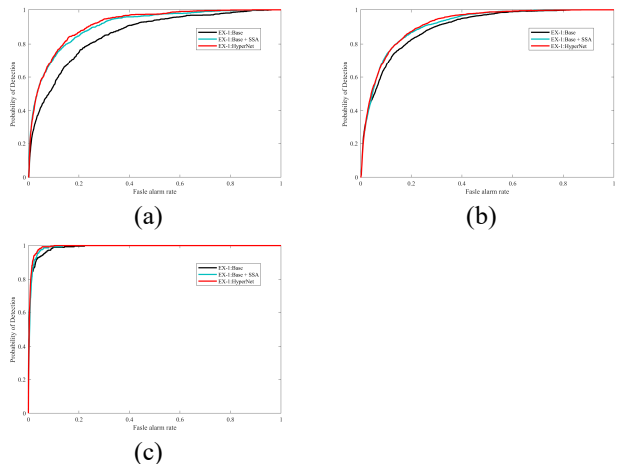

(a)      (b)

(c)

**Fig. 15** The ROC comparison of ablation experiments of HyperNet on three hyperspectral anomalous change detection datasets of (a) EX1: D1F12H1-D1F12H2, (b) EX-2: D1F12H1-D2F22H2, (c) Simulated Hymap dataset.

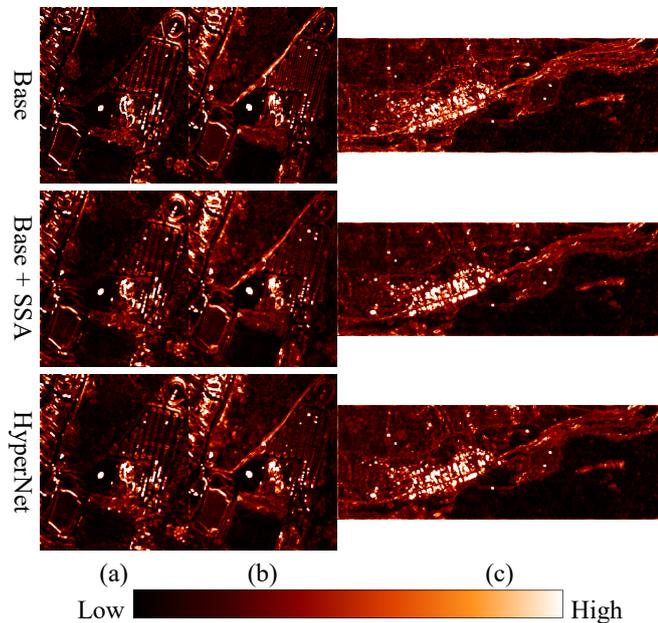

(a)      (b)      (c)

Low ▬▬▬▬▬ High

**Fig. 16** The anomalous change detection map comparison of ablation experiments of (a) EX1: D1F12H1-D1F12H2, (b) EX-2: D1F12H1-D2F22H2, (c) Simulated Hymap dataset. From top to bottom are result from Base, Base + SSA, and HyperNet.

emphasis on hard positive samples promoting the training of network than cosine similarity loss function does.

Moreover, the qualitative and quantitative evaluation of ablation experiments on hyperspectral binary change detection

TABLE VII
THE QUANTITATIVE COMPARISON OF HYPERNET WITH
AND WITHOUT PROPOSED FOCAL COSINE LOSS ON THREE
HYPERSEPCTRAL BINARY CHANGE DETECTION DATASETS

| Dataset | Method | OA | Kappa | F1 | Precision | Recall |
|---|---|---|---|---|---|---|
| Hermiston | Base | 0.9140 | 0.7427 | 0.7969 | 0.8530 | 0.7489 |
| | Base + SSA | 0.9197 | 0.7579 | 0.8083 | **0.8751** | 0.7516 |
| | HyperNet | **0.9206** | **0.7613** | **0.8112** | 0.8740 | **0.7569** |
| Bay | Base | 0.8605 | 0.7211 | 0.8649 | 0.8964 | 0.8355 |
| | Base + SSA | 0.9075 | 0.8144 | 0.9123 | **0.9234** | 0.9016 |
| | HyperNet | **0.9079** | **0.8152** | **0.9129** | 0.9224 | **0.9037** |
| Barbara | Base | 0.8906 | 0.7730 | 0.8647 | 0.8441 | 0.8867 |
| | Base + SSA | **0.9155** | **0.8234** | **0.8934** | **0.8869** | **0.9001** |
| | HyperNet | 0.9114 | 0.8148 | 0.8880 | 0.8836 | 0.8925 |

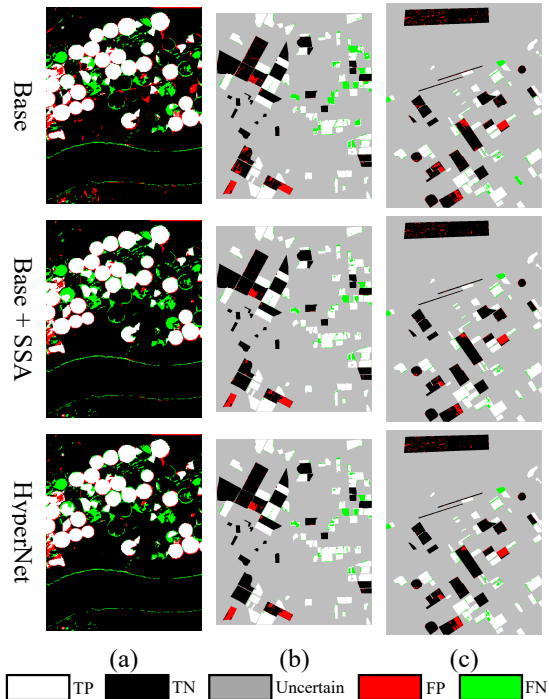

(a)      (b)      (c)

☐ TP ■ TN ▨ Uncertain ■ FP ■ FN

**Fig. 17** The comparison of binary change detection maps of ablation experiments. (a) Hermiston dataset, (b) Bay dataset, (c) Santa Barbara dataset.

have been implemented on another three datasets. As presented in TABLE VII, with the spatial-spectral attention module, "Base + SSA" gain higher OA, Kappa, F1 score, Precision, and recall rate than "Base" does for all three datasets. And it is validated that less false alarm in red and less missing change in green can be found in the detection maps of "Base + SSA" (shown in the second row of Fig. 17) than that of "Base" (shown in the first row of Fig. 17). Besides, HyperNet with cosine loss function gains better OA, Kappa, F1 score, and recall rate than "Base + SSA" does for Hermiston and Bay datasets. However, HyperNet with cosine loss function gains better result on Barbara dataset than HyperNet does, where a little bit more false alarm in red can be found in the result of HyperNet (presented in third row of Fig. 17). The reason may be the hard-positive samples are in a low percentage and the focal cosine loss function plays a less important role compared with the case of Hermiston and Bay datasets, since the "Base" gains higher comprehensive Kappa as 0.7730 for Barbara dataset than for the Hermiston as 0.7427 and Bay as 0.7211.

Overall, HyperNet is capable of obtaining better change detection results, where the spatial-spectral attention module makes a difference in fully exploitation of the spatial relationship and the discriminative spectral features, and the focal cosine plays an important role in amplifying the attention on area with complex spectral difference.

## V. CONCLUSIONS

In this research, we have attempted to put forward a self-supervised spatial-spectral understanding network for hyperspectral change detection. HyperNet employs a self-supervised learning mode and accomplishes multi-temporal spatial and spectral feature comparison in fully convolutional way for pixel-wise feature representation learning. The designed spatial attention branch focuses on the spatial correlation from 2D imaging space, while the spectral attention branch concentrates on the discriminative spectral features for various ground surface objects. And a novel similarity loss function named focal cosine is created for better adjustment of the imbalanced easy and hard positive samples in SSL training. The performance of extensive experiments on six hyperspectral datasets demonstrate the effectiveness and generalization of proposed HyperNet on both HACD and HBCD tasks. In addition, it is proved that the designed spatial-spectral attention branch and focal cosine loss function are capable of detecting more anomalous changes for HACD, reducing the false detection and improve the detection rate for HBCD.